\pdfoutput=1

\documentclass[11pt]{article}

\usepackage[table,xcdraw]{xcolor}

\usepackage{acl}

\usepackage{times}
\usepackage{latexsym}

\usepackage{enumitem}
\usepackage{booktabs}
\usepackage{array}
\usepackage{hyperref}
\usepackage{url}
\usepackage{multirow}
\usepackage{colortbl}
\usepackage{booktabs}
\usepackage{amssymb}
\usepackage{amsmath}
\usepackage{amsthm}
\usepackage{graphicx}
\usepackage{makecell}
\usepackage{threeparttable}
\usepackage{CJKutf8}
\usepackage{lscape}
\usepackage{fancyhdr}

\usepackage[T1]{fontenc}

\usepackage[utf8]{inputenc}

\usepackage{microtype}

\title{MINER: Improving Out-of-Vocabulary Named Entity Recognition from an Information Theoretic Perspective}

\author{{\normalsize
    Xiao Wang$^{\bigstar}$, \ \ Shihan Dou$^{\bigstar}$, \ \ Limao Xiong$^{\bigstar}$, \ \ Yicheng Zou$^{\bigstar}$, } \\ 
    {\normalsize \textbf{Qi Zhang}$^{\bigstar,\clubsuit}$\thanks{{ }{ }Corresponding authors.}\textbf{,} \ \ \textbf{Tao Gui}$^{\blacklozenge}$\textbf{,} \ \
    \textbf{Liang Qiao}$^{\spadesuit}$\textbf{,} \ \ \textbf{Zhanzhan Cheng}$^{\spadesuit}$\textbf{,}\ \ \textbf{Xuanjing Huang}$^{\bigstar}$ } \\
  {$^\bigstar$ \normalsize School of Computer Science, Fudan University, Shanghai, China} \\
  {$^\blacklozenge$ \normalsize Institute of Modern Languages and Linguistics, Fudan University, Shanghai, China} \\
  {$^\clubsuit$ \normalsize Shanghai Key Laboratory of Intelligent Information Processing, Shanghai, China} \\
  {$^\spadesuit$ \normalsize Hikvision Research Institute, Hangzhou, China} \\
  \texttt{\normalsize \{xiao\_wang20,yczou18,qz,tgui,xjhuang\}@fudan.edu.cn}\\
  \texttt{\normalsize \{shdou21,lmxiong21\}@m.fudan.edu.cn} \\
  }

\begin{document}
\maketitle

\begin{abstract}
NER model has achieved promising performance on standard NER benchmarks. However, recent studies show that previous approaches may over-rely on entity mention information, resulting in poor performance on out-of-vocabulary (OOV) entity recognition.  In this work, we propose MINER, a novel NER learning framework, to remedy this issue from an information-theoretic perspective. The proposed approach contains two mutual information-based training objectives: i) generalizing information maximization, which enhances representation via deep understanding of context and entity surface forms; ii) superfluous information minimization, which discourages representation from rote memorizing entity names or exploiting biased cues in data. Experiments on various settings and datasets demonstrate that it achieves better performance in predicting OOV entities.

\end{abstract}

\section{Introduction}

Named Entity Recognition (NER) aims to identify and classify entity mentions from unstructured text, e.g., extracting location mention "Berlin" from the sentence "Berlin is wonderful in the winter". NER is a key component in information retrieval \cite{tan-etal-2021-extracting}, question answering \cite{min-etal-2021-joint}, dialog systems \cite{wang-etal-2020-multi-domain}, etc. Traditional NER models are feature-engineering and machine learning based \cite{zhou2002named,takeuchi2002use,agerri2016robust}. Benefiting from the development of deep learning, neural-network-based NER models have achieved state-of-the-art results on several public benchmarks \cite{lample-etal-2016-neural,peters-etal-2018-deep,devlin2018bert,yamada-etal-2020-luke,yan-etal-2021-unified-generative}.

Recent studies \cite{lin-etal-2020-rigorous,agarwal2021interpretability} show that, context does influence predictions of NER models, but the main factor driving high performance is learning the named tokens themselves. Consequently, NER models underperform when predicting entities that have not been seen during training  \cite{fu2020rethinking,lin-etal-2020-rigorous}, which is referred to as an Out-of-Vocabulary (OOV) problem. 

\begin{table}[t]
\centering
\small
\renewcommand\arraystretch{1.2}
\setlength{\tabcolsep}{3.5pt}{\begin{tabular}{c|ccc||ccc}
\hline
\multirow{2}{*}{} & \multicolumn{3}{c||}{Precision}                        & \multicolumn{3}{c}{Recall}                            \\ \cline{2-7} 
                  & InDict & \multicolumn{1}{c|}{OutDict} & \textbf{Diff} & InDict & \multicolumn{1}{c|}{OutDict} & \textbf{Diff} \\ \hline
PER               & 88.03  & \multicolumn{1}{c|}{75.40}   & 14\%          & 92.90  & \multicolumn{1}{c|}{85.20}   & 8\%           \\ \hline
ORG               & 73.51  & \multicolumn{1}{c|}{72.77}   & 1\%           & 81.93  & \multicolumn{1}{c|}{76.56}   & 7\%           \\ \hline
GPE               & 79.55  & \multicolumn{1}{c|}{78.21}   & 2\%           & 85.37  & \multicolumn{1}{c|}{77.22}   & 10\%          \\ \hline
FAC               & 65.91  & \multicolumn{1}{c|}{65.67}   & 0\%           & 86.05  & \multicolumn{1}{c|}{65.67}   & 24\%          \\ \hline
ALL               & 83.37  & \multicolumn{1}{c|}{71.97}   & \textbf{12\%} & 89.08  & \multicolumn{1}{c|}{79.11}   & \textbf{11\%} \\ \hline
\end{tabular}}
\caption{The comparison between the in-dictionary and out-of-dictionary parts of the CoNLL 2003 baseline \cite{lin-etal-2020-rigorous}, which was tested on Bert-CRF. It is obvious that the performance gap between InDict and OutDict is significantly large.}
\label{tab:result}
\end{table}

There are three classical strategies to alleviate the OOV problem: external knowledge, OOV word embedding, and contextualized embedding. The first one is to introduce additional features, e.g., entity lexicons \cite{zhang-yang-2018-chinese}, part-of-speech tags \cite{li-etal-2018-self}, which alleviates the model's dependence on word embeddings. However, the external knowledge is not always easy to obtain. The second strategy is to get a better OOV word embedding \cite{peng2019learning,fukuda-etal-2020-robust}. The strategy is learning a static OOV embedding representation, but not directly utilizing the context. Last one is fine-tune pre-trained models,  e.g., ELMo \cite{peters-etal-2018-deep}, BERT \cite{devlin2018bert}, which provide contextualized word representations. Unfortunately, \citet{agarwal2021interpretability} shows that the higher performance of pre-trained models could be the results of learning the subword structure better. 

How do we make the model focus on contextual information to tackle the OOV problem? Motivated by the information bottleneck principle \cite{tishby2000information}, we propose a novel learning framework - Mutual Information based Named Entity Recognition (MINER). The proposed method provides an information-theoretic perspective to the OOV problem by training an encoder to minimize task-irrelevant nuisances while keeping predictive information.

Specifically, MINER contains two mutual information based learning objectives: i) generalizing information maximization, which aims to maximize the mutual information between representations and well-generalizing features, i.e., context and entity surface forms; ii) superfluous information minimization, which prevents the model from rote memorizing the entity names or exploiting biased cues via eliminating entity name information. \textit{Our codes}\footnote{https://github.com/BeyonderXX/MINER} are publicly available.

Our main contributions are summarized as follows:

1. We propose a novel learning framework, i.e., MINER, from an information theory perspective, aiming to improve the robustness of entity changes by eliminating entity-specific and maximizing well-generalizing information.

2. We show its effectiveness on several settings and benchmarks, and suggest that MINER is a reliable approach to better OOV entity recognition.

\section{Background}
In this section, we highlight the information bottleneck principle. Subsequently, the analysis of possible issues was provided when applying it to OOV entity recognition. Furthermore, we review related techniques in deriving our framework.

\paragraph{Information Bottleneck}(IB) principle originated in information theory, and provides a theoretical framework for analyzing deep neural networks.  It formulates the goal of representation learning as an information trade-off between predictive power and representation compression. Given the input dataset (X,Y), it seeks to learn the internal representation Z of some intermediate layers by:
$$
L_{IB} = -I(Z;Y) + \beta\ *\ I(Z;X),
$$
where $I$ represents the mutual information(MI), a measure of the mutual dependence between the two variables. The trade-off between the two MI terms is controlled by the Lagrange multiplier $\beta$. A low loss indicates that representation Z does not keep too much information from X while still retaining enough information to predict Y.

Section \ref{Experiments} suggests that directly applying IB to NER can not bring obvious improvement. We argue that IB  cannot guarantee well-generalizing representation.

On the one hand, it has been shown that it is challenging to find a trade-off between high compression and high predictive power \cite{tishby2000information,wang2019deep,piran2020dual}. When compressing task-irrelevant nuisances, however, useful information will inevitably be left out. On the other hand, it is unclear for the IB principle which parts of features are well-generalizing and which are not, as we usually train a classifier to solely maximize accuracy. Consequently, neural networks tend to use any accessible signal to do so \cite{ilyas2019adversarial}, which is referred to as a \textit{shortcut learning} problem \cite{geirhos2020shortcut}. For training sets with limited size, it may be easier for neural networks to memorize entity names rather than to classify them by context and common entity features \cite{agarwal2021interpretability}. In Section \ref{MINER}, we demonstrate how we extend IB to the NER task and address these issues.

\section{Model Architecture}
In recent years, NER systems have undergone a paradigm shift from sequence labeling, which formulates NER as a token-level tagging task \cite{chiu2016named,akbik2018contextual,yan2019tener}, to span prediction (SpanNER), which regards NER as a span-level classification task \cite{mengge-etal-2020-coarse,yamada-etal-2020-luke,fu-etal-2021-spanner}. We choose SpanNER as base architecture for two reasons:

1) SpanNER can yield the whole span representation, which can be directly used for optimize information. 
2) Compared with sequence labeling, SpanNER does better in sentences with more OOV words \cite{fu-etal-2021-spanner}. 

Overall, SpanNER consists of three major modules: token representation layer, span representation layer, and span classification layer. Besides, our method inserts a bottleneck layer to the architecture for information optimization. 

\subsection{Token Representation Layer}
Let $X=\{x_1, x_2, \cdots, x_n\}$ represents the input sentence, thus, the token representation $h_i$ is as follows:

\begin{equation}
\setlength{\abovedisplayskip}{1pt}
\setlength{\belowdisplayskip}{1pt}
u_1, \cdots, u_n = Embedding(x_1, \cdots, x_n)
\end{equation}
\begin{equation}
\setlength{\abovedisplayskip}{1pt}
\setlength{\belowdisplayskip}{1pt}
h_1, \cdots, h_n = Encoder(u_1, \cdots, u_n)
\end{equation}

\begin{figure*}[t]
\centering
  \includegraphics[width=6in]{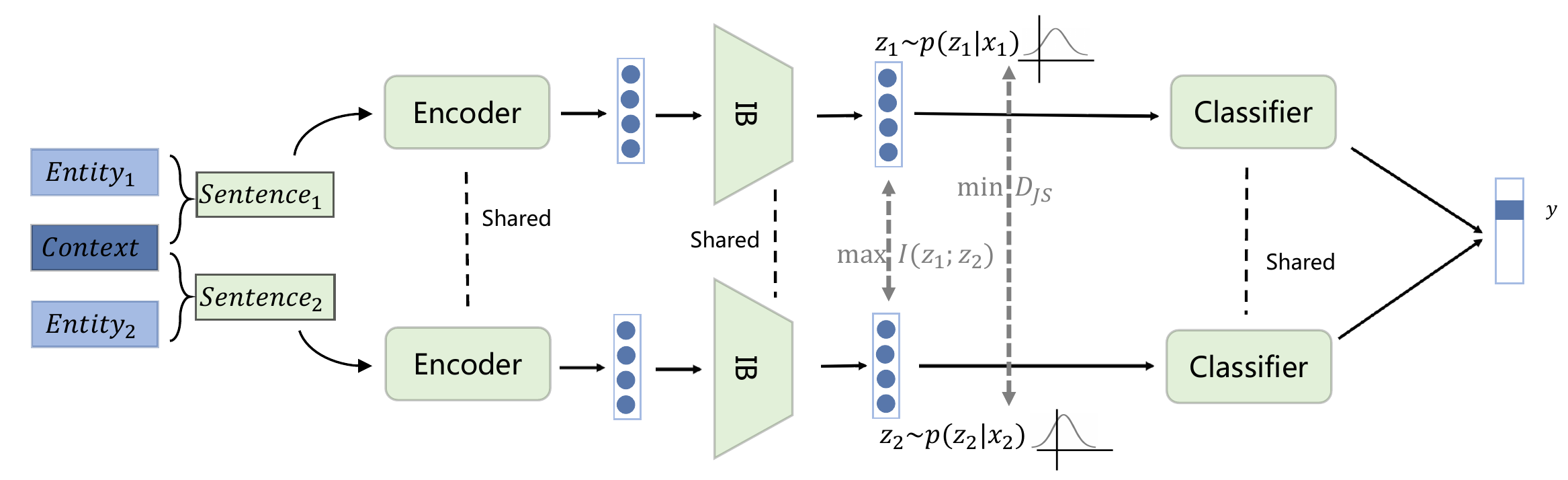}
  \caption{Visualization of MINER, where $x_1$ and $x_2$ share the same context and entity labels, while their entity words are different. $z_1$ and $z_2$ are compressed entity representations sampled by $p(z_1|x_1)$ and $p(z_2|x_2)$, respectively, which are implemented by information bottleneck(IB) layer. Our method add two additional learning objectives to basic architecture. The first one is to maximize the mutual information, i.e., $I(z_1;z_2)$, to enhance context information and entity surface form information of $z_1$ and $z_2$. The second objective is to minimize the Jensen-Shannon divergence, representing an upper bound of $I(x_1;z_1|x_2)$, aiming to eliminate task-irrelevant nuisances.}
 \label{fig:trans}
\end{figure*}

where $Embedding()$ is the non-contextualized word embeddings, e.g., Glove \cite{pennington-etal-2014-glove} or contextualized word embeddings, e.g., ELMo \cite{peters-etal-2018-deep}, BERT \cite{devlin2018bert}. $Encoder()$ can be any network structures with context encoding function, e.g., LSTM \cite{hochreiter1997long}, CNN \cite{lecun1995convolutional}, transformer \cite{vaswani2017attention}, and so on. 

\subsection{Span Representation Layer}
For all possible spans $S=\{s_1, s_2,\cdots,s_m\}$ of sentence $X$, we re-assign a label $y\in Y$ for each span. Take "Berlin is wonderful" as an example, its possible spans and labels are $\{(1,1),(1,2),(1,3),(2,2),(2,3),(3,3)\}$ and $\{LOC,O,O,O,O,O\}$, respectively.

Given the start index $b_i$ and end index $e_i$, the representation of span $s_i$ can be calculated by two parts: boundary embedding and span length embedding.

\textbf{Boundary embedding}: This part is calculated by concatenating the start and end tokens' representation $t_i^b = [h_{b_i}; h_{e_i}]$.

\textbf{Span length embedding}: In order to introduce the length feature, we additionally provide the length embedding $t_i^l$, which can be obtained by a learnable look-up table.

Finally, the span representation can be obtained as: $t_i=[t_i^b;t_i^l]$.

\subsection{Information Bottleneck Layer}
In order to optimize the information in the span representation, our method additionally adds an information bottleneck layer of the form:
\begin{equation}
    p(z|t) = \mathcal{N}\left(z \mid f_{e}^{\mu}(t), f_{e}^{\Sigma}(t)\right)
\end{equation}
where $f_e$ is an MLP which outputs both the K-dimensional mean $\mu$ of $z$ as well as the $K * K$ covariance matrix $\Sigma$. Then we 
can use the reparameterization trick (\cite{kingma2013auto}) to get the compressed representation $z_i$.

\subsection{Span Classification Layer}
Once the information bottleneck layer is finished, $z_i$ is fed into the classifier to obtain the probability of its label $y_i$. Based on the probability, the basic loss function can be calculated as follows:
\begin{equation}
    L_{base} = - \frac{score(z_i, y_i)}{\sum_{y^{'} \in Y}score(z_i, y')},
    \label{origin loss}
\end{equation}
where $score()$ is a function that measures the compatibility between a specified label and a span representation:
\begin{equation}
    score(z_i,y^k)=exp(z_i^{T}y^k),
\end{equation}
where $y^k$ is a learnable representation of class k.

\textbf{Heuristic Decoding}  A heuristic decoding solution for the flat NER is provided to avoid the prediction of over-lapped spans. For those overlapped spans, we keep the span with the highest prediction probability and drop the others.

It's worth noting that our method is flexible and can be used with any other NER model based on span classification. In next section, we will introduce two additional objectives to tackle the OOV problem of NER.

\section{MI-based objectives}\label{MINER}
Motivated by IB \cite{tishby2000information,federici2020learning}, we can subdivide $I(X;Z)$ into two components by using the chain rule of mutual information(MI):

\begin{equation}
    I(X;Z) = \underbrace{I(Y;Z)}_{predictive} + \underbrace{I(X;Z|Y)}_{superfluous},
    \label{origin_target}
\end{equation}

The first term determines how much information about Y is accessible from Z. While the second term, conditional mutual information term $I(X;Z|Y)$, denotes the information in $Z$ that is not predictive of $Y$. 

\textit{For NER, which parts of the information retrieved from input  are useful and which are redundant?}

From human intuition, \textbf{text context} should be the main predictive information for NER. For example, "The CEO of $X$ resigned", the type of $X$ in each of these contexts should always be "ORG". Besides, \textbf{entity mentions} also provide much information for entity recognition. For example, nearly all person names capitalize the first letter and follow the "firstName lastName" or "lastName firstName" patterns. 
However, \textbf{entity name} is not a well-generalizing features. By simply memorizing the fact which span is an entity, it may be possible for it to fit the training set, but it is impossible to predict entities that have never been seen before.


We convert the targets of Eq. (\ref{origin_target}) into a form that is easier to solve via a contrastive strategy. 
Specifically, consider $x_1$ and $x_2$ are two contrastive samples of similar context, and contains different entity mentions of the same entity category, i.e., $s_1$ and $s_2$, respectively. Assuming both $x_1$ and $x_2$ are both \textbf{sufficient} for inferring label $y$. The mutual information between $x_1$ and $z_1$ can be factorized to two parts.

\begin{equation}
\setlength{\abovedisplayskip}{1pt}
\setlength{\belowdisplayskip}{10pt}
I(x_1;z_1) = \underbrace{I(z_1;x_2)}_{consistent} + \underbrace{I(x_1;z_1|x_2)}_{specific}, 
\label{approximation}
\end{equation}
where $z_1$ and $z_2$ are span representations of $s_1$ and $s_2$, respectively, $I(z_1; x_2)$ denotes the information that isn't entity-specific. And $I(x_1;z_1|x_2)$ represents the information in $z_1$ which is unique to $x_1$ but is not predictable by sentence $x_2$, i.e., entity-specific information.

Thus any representation $z$ containing all information shared from both sentences would also contain the necessary label information, and sentence-specific information is superfluous. So Eq. (\ref{origin_target}) can be approximated by Eq. (\ref{approximation}) by:
\begin{equation}
\setlength{\abovedisplayskip}{10pt}
\setlength{\belowdisplayskip}{1pt}
maximize\ I(z_1; y)\  \sim \  I(z_1; x_2),
\label{GIM}
\end{equation}
\begin{equation}
\setlength{\abovedisplayskip}{1pt}
\setlength{\belowdisplayskip}{10pt}
minimize\ I(x_1;z_1| y) \ \sim \ I(x_1;z_1|x_2), 
\label{SIM}
\end{equation}

The target of Eq. (\ref{GIM}) is defined as \textbf{generalizing information} maximization. We proved that $I(z_1; z_2)$ is a lower bound of $I(z_1;x_2)$(proof could be found in appendix \ref{appendix}). InfoNCE \cite{oord2018representation} was used as a lower bound on MI and can be used to approximate $I(z_1; z_2)$. Subsequently, it can be  optimized by:

\begin{equation}
\small L_{gi} = - \mathbb{E}_p \left [ g_w(z_1,z_2) - \mathbb{E}_{p^\prime} {\log{\sum_{z^{\prime}} {\exp g_w(z_1,z^{\prime})}}} \right ],
\label{MI}
\end{equation} 

where $g_w(\cdot,\cdot)$ is a compatible score function approximated by a neural network, $z_2$ are the positive entity representations from the joint distribution $p$ of original sample and corresponding generated sample, $z^\prime$ are the negative entity representations drawn from the joint distribution of the original sample and other samples.

The target of Eq. (\ref{SIM}) is defined as \textbf{superfluous information} minimization. To restrict this term, we can minimize an upper bound of $I(x_1;z_1|x_2)$ (proofs could be found in appendix \ref{appendix}) as follows:

\begin{equation}
L_{si} = \mathbb{E}_{x_1,x_2} \mathbb{E}_{z_1,z_2} \left [ D_{JS} {\left [ {p_{z_1}||p_{z_2}} \right ]}  \right ] ,
\label{OOV}
\end{equation}

where $D_{JS}$ means Jensen-Shannon divergence, $p_{z_1}$ and $p_{z_2}$ represent $p(z_1|x_1)$ and $p(z_2|x_2)$, respectively. In practice, Eq. (\ref{OOV}) encourage $z$ to be invariant to entity changes.
The resulting Mutual Information based Named Entity Recognition model is visualized in Figure \ref{fig:trans}.

 \subsection{Contrastive sample generation}
It is difficult to obtain samples with similar contexts but different entity words. We generate contrastive samples by the mention replacement mechanism\cite{dai2020analysis}. For each mention in the sentence, we replace it by another mention from the original training set, which has the same entity type. The corresponding span label can be changed accordingly. For example, "LOC" mention "Berlin" in sentence "Berlin is wonderful in the winter" is replaced by "Iceland". 

\subsection{Training}
Combine Eq. (\ref{origin loss}), (\ref{MI}), and (\ref{OOV}), we can get the following objective function, which try to minimize:

\begin{equation}
    L = L_{base} + \gamma * L_{gi} + \beta * L_{si},
\end{equation} 

where $\gamma$ and $\beta$ are the weights of the generalizing information loss and superfluous information loss, respectively.

\section{Experiment}\label{Experiments}
In this section, we verify the performance of the proposed method on five OOV datasets, and compared it with other methods. In addition, We tested the universality of the proposed method in various pre-trained models.

\begin{table}[t]
\begin{tabular}{lccc}
\hline
\hline
Datasets        & \multicolumn{1}{l}{sents} & \multicolumn{1}{l}{entities} & \multicolumn{1}{l}{OOV Rate} \\ \hline
WNUT2017        & 1286               &  947                   & 1.00                            \\ \hline
TwitterNER      & 3257              & 3990                  & 0.62                   \\ \hline
BioNER          & 3856              &  4344                 & 0.77                      \\ \hline
Conll2003-Typos & 2676                      & 4130                         & 0.71                      \\ \hline
Conll2003-OOV   & 3684                      & 5648                         & 0.96                      \\ \hline

\end{tabular}
\caption{Number of OOV entities in the test sets. }
\label{dataset}
\end{table}

\begin{table*}[h]
\renewcommand\arraystretch{1.2}
\centering
\setlength{\tabcolsep}{8.5pt}{\begin{tabular}{c|c|c|c|c|c}
\hline
\hline
& & & & \multicolumn{2}{c}{\textbf{CoNLL 2003}} \\\cline{5-6}
\multirow{-2}{*}{\textbf{Methods}} &
\multirow{-2}{*}{\textbf{WNUT2017}} & 
\multirow{-2}{*}{\textbf{BioNER}} & 
\multirow{-2}{*}{\textbf{TwitterNER}} & \multicolumn{1}{c|}{Typos}&
\multicolumn{1}{c}{OOV}            \\ \hline 

\textbf{VaniIB} &51.60 &73.41 &71.19 & \multicolumn{1}{c|}{83.49} & 70.12 \\ \hline
\textbf{DataAug} &52.29 &75.85 &73.69 
&\multicolumn{1}{c|}{81.73} &69.6 \\ \hline
\textbf{InferNER} & 50.52 & - &74.17 & \multicolumn{1}{c|}{-} & - \\ \hline
\textbf{MIN}  &49.93 & \textbf{77.97}  & - 
&\multicolumn{1}{c|}{-} & - \\ \hline
\textbf{CoFEE} &39.1 & - &69.5                                             & \multicolumn{1}{c|}{-} & -              \\ \hline
\textbf{MAML}  & 24.19 & 76.36 & -                                         & \multicolumn{1}{c|}{-} & -              \\ \hline
\textbf{SA-NER}   & 50.36   & - & -                                        & \multicolumn{1}{c|}{-}  & -               \\ \hline
\textbf{SpanNER (Bert large)} &51.83 &73.78  &71.57                        &\multicolumn{1}{c|}{81.83} &64.43          \\ \hline
\textbf{SpanNER (Roberta large)} &51.65 &74.49 &71.7                       & \multicolumn{1}{c|}{82.85} &64.7           \\ \hline
\textbf{SpanNER (AlBert large)} &49.13 &71.08 &70.33 
&\multicolumn{1}{c|}{82.49} &64.12   \\ \hline

\textbf{MINER (Bert large)}  & 54.52 & 77.03 & 75.26                  & \multicolumn{1}{c|}{87.09} & 78.03          \\ \hline
\textbf{MINER (Roberta large)} & \textbf{54.86} & 76.43               & \textbf{75.38}   
& \multicolumn{1}{c|}{\textbf{87.57}}          & \textbf{79.15} \\ \hline
\textbf{MINER (Albert large)}  & 51.94 & 75.23 & 72.67                & \multicolumn{1}{c|}{86.53} & 77.95          \\ \hline
\hline
\end{tabular}}
\caption{Performance of the proposed method compared with state-of-the-arts.}
\label{tab:performance}
\end{table*}

\subsection{Datasets and Metrics}
\paragraph{Datasets} We performed experiments on: 

\begin{enumerate}
\item WNUT2017 \cite{derczynski2017results}, a dataset focus on unusual, previous-unseen entities in training data, and is collected from social media.

\item TwitterNER \cite{zhang2018adaptive}, an English NER dataset created from Tweets.

\item BioNER \cite{kim2004introduction}, the JNLPBA 2004 Bio-NER dataset focus on technical terms in the biology domain.

\item Conll03-Typos \cite{wang2021textflint}, which is generated from Conll2003 \cite{sang2003introduction}. The entities in the test set are replaced by typos version(character modify, insert, and delete operation).

\item Conll03-OOV \cite{wang2021textflint}, which is generated from Conll2003 \cite{sang2003introduction}. The entities in the test set are replaced by another out-of-vocabulary entity in test set.

\end{enumerate}

Table \ref{dataset} reports the statistic results of the OOV problem on the test sets of each dataset. As shown in the table, the test set of these datasets comprises a substantial amount of OOV entities.

\paragraph{Metrics} We measured the entity-level micro average F1 score on the test set to compare the results of different models.

\subsection{Baseline methods}
\citet{li-etal-2020-handling} share the same intuition as us, enriching word representations with context. However, the work is neither open source nor reported on the same dataset, so this method cannot be compared with MINER. We compare our method with baselines as follows:

\begin{itemize}
\item \citet{fu-etal-2021-spanner} (SpanNER), which is trained by original SpanNER framework, without any constraint and extra data processing.  

\item Vanilla information bottleneck(VaniIB), a method employs the original information bottleneck constraint to the SpanNER, which is optimized based on \citet{alemi2016deep}. Compared with our method, it directly compresses all the information from the input.

\item \citet{dai2020analysis} (DataAug) , which trains model with data augmentation strategy, while keeps the same model architecture as SpanNER. This model is trained by 1:1 original training set and entity replacement training set, which keeps the same input as the proposed method.

\item \citet{shahzad2021inferner} (InferNER), a method focus on word-, character-, and sentence-level information for NER in short-text, without recurring to external sources. In addition, it is able to incorporate visual information and introduce an attention component which computes attention weight probabilities over textual and text-relevant visual contexts separately. 

\item \citet{li-etal-2021-modularized} (MIN), which utilizes both segment-level information and word-level dependencies, and incorporates an interaction mechanism to support information sharing between boundary detection and type prediction, enhancing the performance for the NER task.

\item \citet{fukuda-etal-2020-robust} (CoFEE), which refer to pre-trained word embeddings for known words with similar surfaces to target OOV words. 

\item \citet{nie2020named} (SA-NER), which utilize semantic enhancement methods to reduce the negative impact of data sparsity problems. Specifically, the method obtains the augmented semantic information from a large-scale corpus, and proposes an attentive semantic augmentation module and a gate module to encode and aggregate such information, respectively.

\end{itemize}

To verify the universality of our method, we measured its performance on various pre-trained models, i.e., Bert \cite{devlin2018bert},  Roberta \cite{liu2019roberta}, Albert \cite{lan2019albert}.

\subsection{Implementation Details}
Bert-large released by \citet{devlin2018bert} is selected as our base encoder. The learning rate is set to 5e-5, and the dropout is set to 0.2. The output dim of the information bottleneck layer is 50. In order to make a trade-off for the performance and efficiency, on the one hand, we truncate the part of the sentence whose tokens exceeds 128. On the other hand, we count the length distribution of entity length in different datasets, and finally choose 4 as the maximum enumerated entity length. The values of $\beta$ and $\gamma$ differ for different datasets. Empirically, 1e-5 for $\beta$ and 0.01 for $\gamma$ can get promised results. The model is trained in an NVIDIA GeForce RTX 2080Ti GPU. Checkpoints with top-3 performance are finally evaluated on the test set to report averaged results.

\subsection{Main Results}
We demonstrate the effectiveness of MINER against other state-of-the-art models. As shown in table \ref{tab:performance}, we conducted the following comparison and analysis:


1) Our baseline model, i.e., SpanNER, does an excellent job of predicting OOV entities. Compared with sequence labeling, the span classification could model the relation of entity tokens directly;2) The performance of SpanNER is further boosted with our proposed approach, which proved the effectiveness of our method. As shown in table \ref{tab:performance}, MINER almost outperforms all other SOTA methods without any external resource;3) Compared with \textit{Typos} data transformation, it is more difficult for models to predict \textit{OOV} words. To pre-trained model, typos word may not appear in training set, but they share most subwords with the original token. Moreover, the subword of OOV entity may be rare; 4) It seems that the traditional information bottleneck will not significantly improve the OOV prediction ability of the model. We argue that the traditional information bottlenecks will indiscriminately compress the information in the representation, leading to underfitting; 5) Our model has significantly improved the performance of the model on the entity perturbed methods of typos and OOV, demonstrating that MI improve the robustness substantially in the face of noise; 
6) It is clear that our proposed method is universal and can further improve OOV prediction performance for different embedding models, as we get improvements on Bert, Roberta, and Albert stably.

\begin{figure}[t]
\small
\centering
  \includegraphics[width=3.0in]{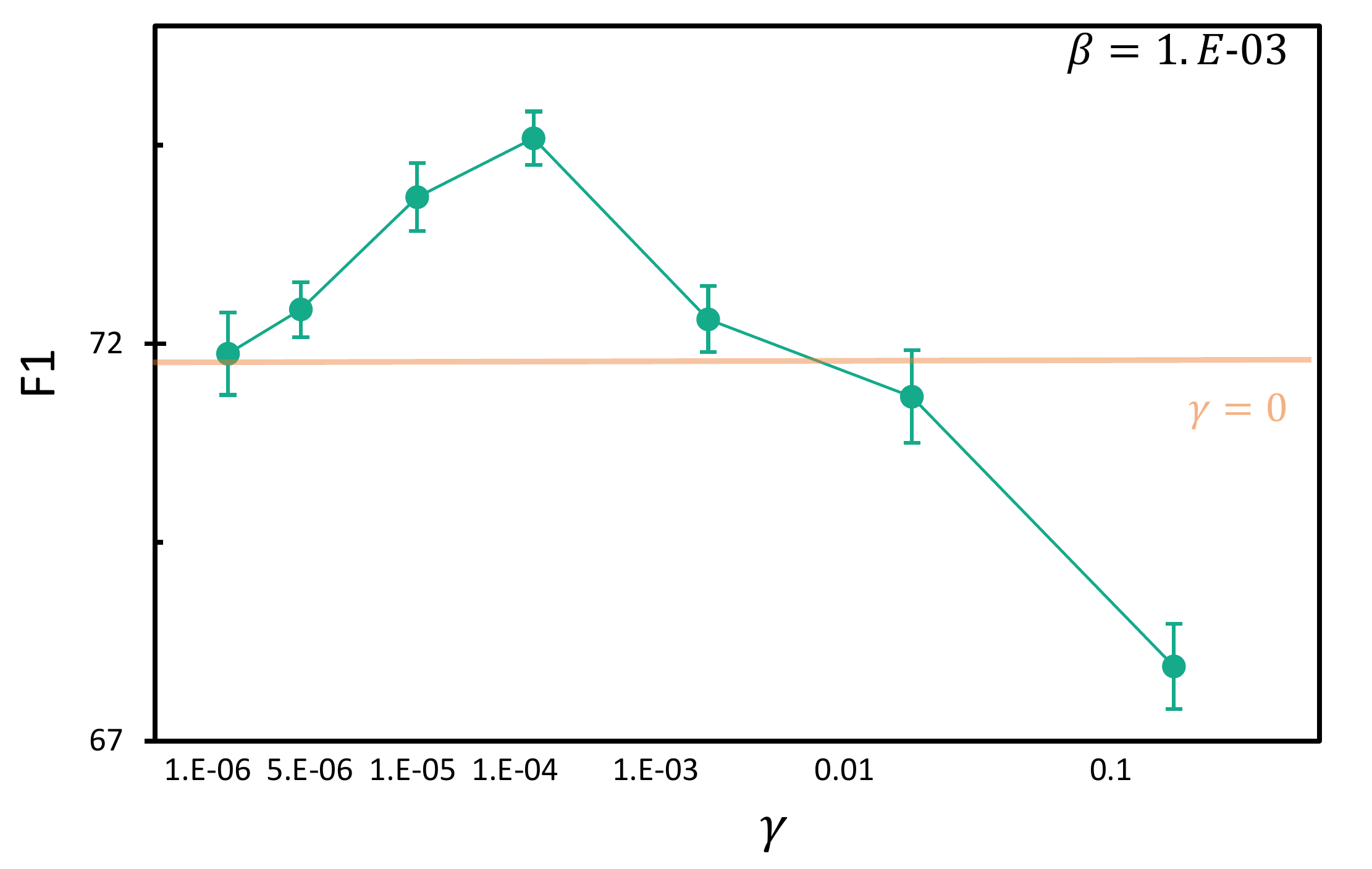}
  \caption{Illustration of f1 score in different $\gamma$ values. The results are obtained by testing MINER (Bert large) on TwitterNER \cite{zhang2018adaptive}. We fix $\beta=1e\-03$, and the orange line is f1 score when $\gamma=0$.}
 \label{sen_analysis}
\end{figure}

\begin{figure}[t]
\small
\centering
  \includegraphics[width=3.0in]{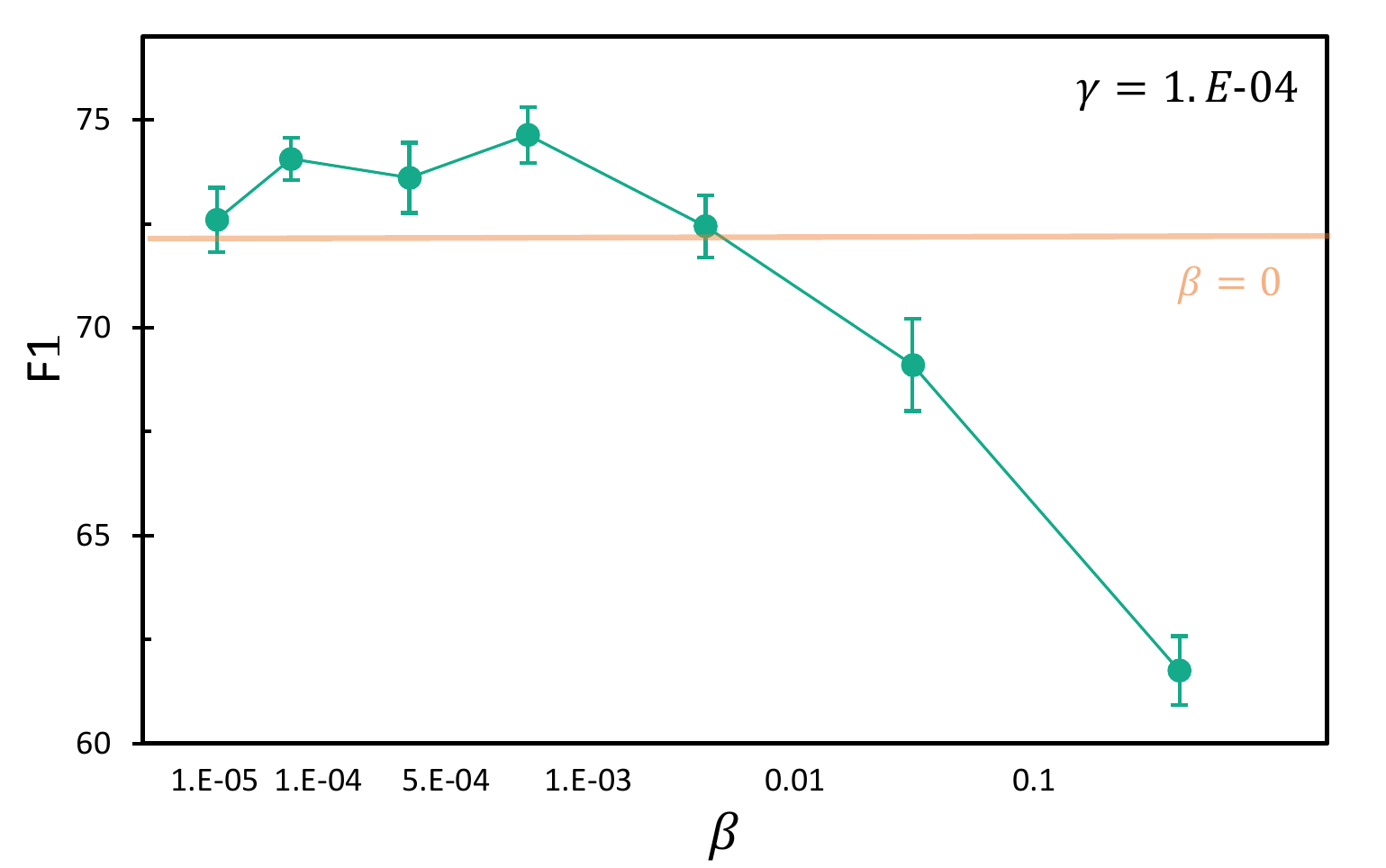}
  \caption{Illustration of f1 score in different $\beta$ values. The results are obtained by testing MINER (Bert large) on TwitterNER \cite{zhang2018adaptive}. We fix $\gamma=1e\-04$, and the orange line is f1 score when $\beta=0$.}
 \label{ablation}
\end{figure}

\begin{figure*}[h]
\small
\centering
  \includegraphics[width=6in]{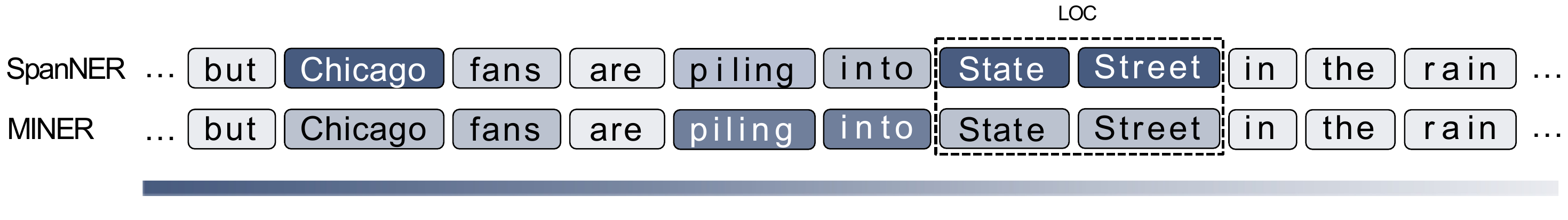}
  \caption{Visualization of attention weights over entities and context.}
 \label{attention}
\end{figure*}

\begin{table}[t]
\footnotesize
\begin{center}
\begin{tabular}{l|ccc}
\toprule[1pt]
\bf Dataset &\bf OOV & \bf MI & \bf F1\\
\midrule
\multirow{4}{*}{{\bf WNUT 2017}} &- &- &51.83\\
& \checkmark&- &52.57\\
&- & \checkmark&53.91\\
&\checkmark &\checkmark &\bf 54.52\\
\midrule
\multirow{4}{*}{{\bf BioNER}} &- &- &73.78\\
&\checkmark &- &75.23\\
&- &\checkmark &74.22\\
&\checkmark &\checkmark &\bf 77.03\\
\midrule
\multirow{4}{*}{{\bf Twitter-NER}} &- &- &71.57\\
&\checkmark &- &73.78\\
&- &\checkmark &73.32\\
&\checkmark &\checkmark &\bf 75.26\\
\bottomrule[1pt]
\end{tabular}
\end{center}
\caption{\label{ablation}Ablation study results on three datasets.}
\end{table}

\subsection{Ablation Study}
We also perform ablation studies to validate the effectiveness of each part in MINER. Table \ref{ablation} demonstrates the results of different settings for the proposed training strategy equipped with BERT. After only adding the $L_{gi}$ loss to enhance context and entity surface form information, we find that the results are better than the original PLMs. A similar phenomenon occurs in $L_{si}$, too. It reflects that both $L_{gi}$ and $L_{si}$ are beneficial to improve the generalizing ability on OOV entities recognition. Moreover, the results on the three datasets are significantly improved by adding both $L_{gi}$ and $L_{si}$ learning objectives. It means $L_{gi}$ and $L_{si}$ can boost each over, which proves that our method enhances representation via deep understanding of context and entity surface forms and discourages representation from rote memorizing entity names or exploiting biased cues in data.

\subsection{Sensitivity Analysis of $\beta$ and $\gamma$}

To show the different influence of our proposed training objectives $L_{gi}$ and $L_{si}$, we conduct sensitivity analysis of the coefficient $\beta$ and $\gamma$. Figure \ref{sen_analysis} shows the performance change under different settings of the two coefficients. The yellow line denotes ablation results without the corresponding loss functions (with $\beta$=0 or $\gamma$=0). From Figure \ref{sen_analysis} we can observe that the performance is significantly enhanced with a small rate of $\beta$ or $\gamma$, where the best performance is achieved when $\beta$=1e-3 and $\gamma$=1e-4, respectively. It probes the effectiveness of our proposed training objectives that enhances representation via deep understanding of context and entity surface forms and discourages representation from rote memorizing entity names or exploiting biased cues in data. As the coefficient rate increases continuously, the performance shows a declining trend, which means the over-constraint of $L_{gi}$ or $L_{si}$ will hurt the generalizing ability of predicting the OOV entities.


\subsection{Interpretable Analysis}
The above experiments show the promising performance of MINER on predicting the unseen entities. To further investigate which part of the sentence MINER focuses on, we visualize the attention weights over entities and contexts. We demonstrate an example in Figure \ref{attention} , where is selected from TwitterNER. The attention score is calculated by averaging the attention weight of the 0th layer of BERT. Take the attention weights of the entity "State Street" as an example, it is obvious that baseline model, i.e., SpanNER, focus on entity words themselves. While the scores of our model are more average, it means that our method concerns more context information.  

\section{Related Work}
\subsection{External Knowledge}
This group of methods makes it easier to predict OOV entities using external knowledge. \citet{zhang-yang-2018-chinese} utilize a dictionary to list numerous entity mentions. It is possible to get stronger "look-up" models by integrating dictionary information, but there is no guarantee that entities outside the training set and vocabulary will be correctly identified. To diminish the model's dependency on OOV embedding, \citet{li-etal-2018-self} introduce part-of-speech tags. External resources are not always available, which is a limitation of this strategy.

\subsection{OOV word Embedding}
The OOV problem can be alleviated by improving the OOV word embedding. The character ngram of each word is used by \citet{bojanowski-etal-2017-enriching} to represent the OOV word embedding. \citet{pinter-etal-2017-mimicking} captures morphological features using character-level RNN. Another technique is to first match the OOV words with the words that have been seen in training, then replace the OOV words' embedding with the seen words' embedding. \citet{peng2019learning} trains a student network to predict the closest word representation to the OOV term. \citet{fukuda-etal-2020-robust} referring to pre-trained word embeddings for known words with similar surfaces to target OOV words. This kind of method is learning a static OOV embedding representation, and does not directly utilize the context.

\subsection{Contextualized Embedding}
Contextual information is used to enhance the representation of OOV words in this strategy. \cite{hu-etal-2019-shot} formulate the OOV problem as a K-shot regression problem and learns to predict the OOV embedding by aggregating only K contexts and morphological features. Pre-trained models contextualized word embeddings via pretraining on large background corpora.
Furthermore, contextualized word embeddings can be provided by the pre-trained models, which are pre-trained on large background corpora \cite{peters-etal-2018-deep,devlin2018bert,liu2019roberta}. \citet{yan-etal-2021-unified-generative} shows that BERT is not always better at capturing context as compared to Gloe-based BiLSTM-CRFs. Their higher performance could be the result of learning the subword structure better.

\section{Conclusion}
Based on the recent studies of NER, we analyze how to improve the OOV entity recognition. In this work, we propose a novel and flexible learning framework - MINER, to tackle OOV entities recognition issue from an information-theoretic perspective. On the one hand, this method can enhance the context information of the output of the encoder. On the other hand, it can safely eliminate task-irrelevant nuisances and prevents the model from rote memorizing the entities. Specifically, the proposed approach contains two mutual information based training objectives: generalizing information maximization, and superfluous information minimization. Experiments on various datasets demonstrate that MINER achieves much better performance in predicting out-of-vocabulary entities.

\section*{Acknowledgements}
The authors would like to thank the anonymous reviewers for their helpful comments, Ting Wu and Yiding Tan for their early contribution. This work was partially funded by China National Key R\&D Program (No. 2018YFB1005104), National Natural Science Foundation of China (No. 62076069, 61976056). This research was sponsored by Hikvision Cooperation Fund, Beijing Academy of Artificial Intelligence(BAAI), and CAAI-Huawei MindSpore Open Fund.

\bibliography{anthology,custom}
\bibliographystyle{acl_natbib}

\appendix
\label{appendix}

\section{Appendix}
\label{sec:appendix}
This section provides the proof of generalizing information maximization, i.e., Eq. (\ref{GIM}). Consider $x_1$ and $x_2$ are two contrastive samples of similar context, and contains different entity mentions of the same entity category, i.e., $s_1$ and $s_2$, respectively. 

\begin{equation}
\begin{aligned}
 I(z_1; x_2) =& I(z_1;x_2z_2) - I(z_1;z_2|x_2)\\
=& I(z_1;x_2z_2)\\
=& I(z_1;z_2) + I(z_1;x_2|z_2)\\
\geq& I(z_1;z_2)
\end{aligned}   
\end{equation}

\section{Appendix}
\label{sec:appendix}
This section provides the proof of superfluous information minimization, i.e. Eq. (\ref{SIM}). 

\begin{equation}
\begin{aligned}
&\ \ \ \ \ I(x_1;z_1|x_2) \vspace{4pt}\\
&= E_{x_1,x_2\sim p(x_1,x_2)}E_{z\sim p(z_1|v_1)} \log \frac{p(x_1, z_1|x_2)}{p(x_1|x_2)p(z_1|x_2)}\vspace{8pt}\\
&= E_{x_1,x_2\sim p(x_1,x_2)}E_{z\sim p(z_1|v_1)} \log \frac{p(z_1|x_1)p(x_1|x_2)}{p(x_1|x_2)p(z_1|x_2)} \vspace{8pt}\\
&= E_{x_1,x_2\sim p(x_1,x_2)}E_{z\sim p(z_1|v_1)} \log \frac{p(z_1|x_1)}{p(z_1|x_2)}\vspace{8pt}\\
&= E_{x_1,x_2\sim p(x_1,x_2)}E_{z\sim p(z_1|v_1)} \log \frac{p(z_1|x_1)p(z_2|x_2)}{p(z_2|x_2)p(z_1|x_2)}\vspace{8pt}\\
&= D_{KL}(p(z_1|x_1) || p(z_2|x_2)) \vspace{6pt} - D_{KL}(p(z_1|x_2) || p(z_2|x_2))\vspace{6pt}\\
&\leq D_{KL}(p(z_1|x_1) || p(z_2|x_2)) \vspace{6pt}
\end{aligned}   
\end{equation}

\end{document}